%% file: fineline.tex
\documentclass{article}

\usepackage{arxiv}

\usepackage[utf8]{inputenc} 
\usepackage[T1]{fontenc}    
\usepackage{hyperref}       
\usepackage{url}            
\usepackage{booktabs}       
\usepackage{amsfonts}       
\usepackage{nicefrac}       
\usepackage{microtype}      
\usepackage[pdftex]{graphicx}
\usepackage{amssymb,amsfonts,amsmath,amscd}
\usepackage[printonlyused,withpage]{acronym}
\usepackage{natbib}
\usepackage{mathrsfs}
\usepackage{rotating}

\graphicspath{{figs/}}
\input{notation_header}

\acrodef{BA}{Bundle Adjustment}
\acrodef{SLAM}{Simultaneous Localization and Mapping}
\acrodef{RANSAC}{Random Sample And Consensus}
\acrodef{QCQP}{Quadratically Constrained Quadratic Program}
\acrodef{SDP}{Semi-Definite Program}
\acrodef{IRLS}{Iteratively Reweighted Least-Squares}
\acrodef{TLS}{Total Least-Squares}
\acrodef{SVD}{Singular-Value Decomposition}
\acrodef{GM}{Geman-McClure}
\acrodef{PSD}{positive-semidefinite}

\hypersetup{%
    pdftitle={},
    pdfauthor={},
    pdfkeywords={},
    pdfsubject={},
    pdfstartview=FitH,%
    bookmarks=true,%
    bookmarksopen=true,%
    breaklinks=true,%
    colorlinks=true,%
    linkcolor=black,
    anchorcolor=black,%
    citecolor=black,
    filecolor=black,%
    menucolor=black,
    pagecolor=black,%
    urlcolor=black
}%

\title{A Fine Line: Total Least-Squares Line Fitting \\ as QCQP Optimization}

\author{
 {\normalfont Timothy D. Barfoot} \\
 Robotics Institute \\
 University of Toronto \\
 \texttt{tim.barfoot@utoronto.ca} 
 \and
 {\normalfont Connor Holmes} \\
 Robotics Institute \\
 University of Toronto \\
 \texttt{connor.holmes@mail.utoronto.ca} 
 \and \\
 {\normalfont Frederike D\"{u}mbgen} \\
 Robotics Institute \\
 University of Toronto \\
 \texttt{frederike.dumbgen@utoronto.ca} 
}

\date{}

\begin{document}

\maketitle
\title{A Fine Line}

\begin{abstract}
This note uses the \ac{TLS} line-fitting problem as a canvas to explore some modern optimization tools.  The contribution is meant to be tutorial in nature.  The \ac{TLS} problem has a lot of mathematical similarities to important problems in robotics and computer vision but is easier to visualize and understand.  We demonstrate how to turn this problem into a \ac{QCQP} so that it can be cast either as an eigenproblem or a \ac{SDP}.  We then turn to the more challenging situation where a Geman-McClure cost function and M-estimation are used to reject outlier datapoints.  Using Black-Rangarajan duality, we show this can also be cast as a \ac{QCQP} and solved as an \ac{SDP}; however, with a lot of data the \ac{SDP} can be slow and as such we show how we can construct a certificate of optimality for a faster method such as \ac{IRLS}.
\end{abstract}


\section{Total Least-Squares}

\subsection{Primal Problem}

\begin{figure}[t]
\centering
\includegraphics[height=0.3\textwidth]{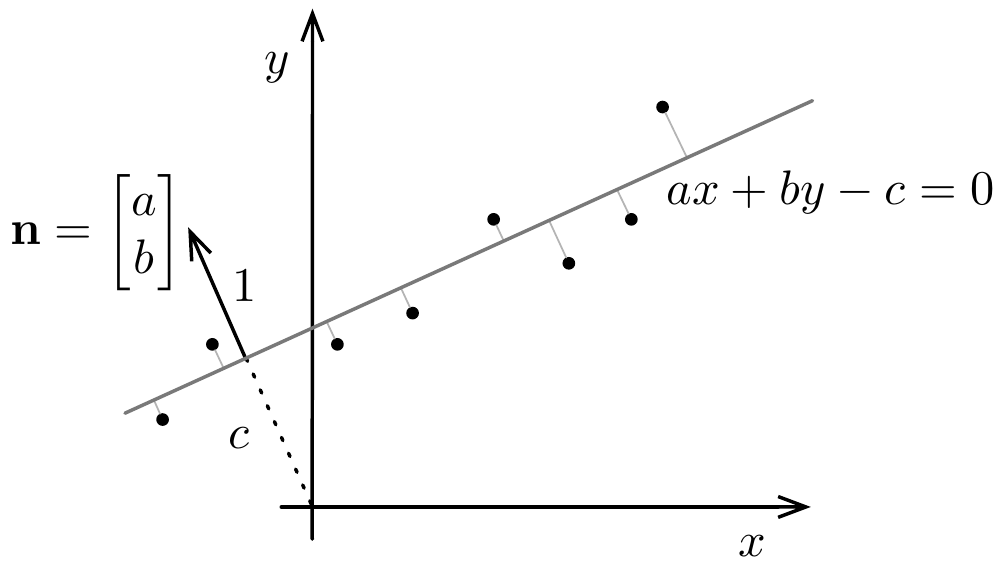}
\caption{The \ac{TLS} problem. The line parameters $a$ and $b$ form the unit-normal for the line.  The parameter $c$ is the offset of the line from the origin.}
\label{fig:lineparam}
\end{figure}

In the \ac{TLS} problem, we are interested in the simple scenario of fitting a straight line, $ax + by - c = 0$, to a collection of noisy data points, $(x_n,y_n)$.  The basic optimization problem that we want to solve is
\begin{equation}\label{eq:tls}
\begin{array}{rc}
\mbox{min} & p(a,b,c) = \frac{1}{N}\sum_n (ax_n + by_n - c)^2 \\
\mbox{w.r.t.} & a,b,c \\
\mbox{s.t.} & a^2 + b^2 = 1
\end{array} ,
\end{equation}
where $N \geq 2$ is the total number of datapoints.
Figure~\ref{fig:lineparam} shows the parameterization of our line and the \ac{TLS} errors graphically.  We see that $c$ is the offset of the line from the origin.  The parameters $a$ and $b$ form a normal to the line and the constraint, $a^2 + b^2 = 1$, comes from the fact that we want it to be a unit normal.  This optimization problem is known as a {\em \acf{QCQP}}, since the cost function is quadratic and there is a quadratic constraint on the design variables.

With a bit of manipulation, we can write the objective function as
\begin{equation}
p(a,b,c) = \underbrace{\bbm a \\ b \ebm^T}_{\mbf{n}^T} \underbrace{\bbm \frac{1}{N} \sum_n x_n^2 & \frac{1}{N} \sum_n x_n y_n \\  \frac{1}{N} \sum_n x_n y_n & \frac{1}{N} \sum_n y_n^2 \ebm}_{\mbf{A}} \underbrace{\bbm a \\ b \ebm}_{\mbf{n}} - 2 c \underbrace{\bbm \frac{1}{N} \sum_n x_n \\ \frac{1}{N} \sum_n y_n \ebm^T}_{\mbf{b}^T}\underbrace{\bbm a \\ b \ebm}_{\mbf{n}} + \; c^2,
\end{equation}
where $\mbf{n}$ is the unit normal.  We see that $\mbf{b}$ is the centroid of our datapoints and $\mbf{A}$ is the covariance with respect to the origin.  Our optimization problem can then be written as
\begin{equation}
\begin{array}{rc}
\mbox{min} & p(\mbf{n},c) = \mbf{n}^T \mbf{A} \mbf{n} - 2 c \, \mbf{b}^T \mbf{n} + c^2 \\
\mbox{w.r.t.} & \mbf{n},c \\
\mbox{s.t.} & \mbf{n}^T \mbf{n} = 1
\end{array} .
\end{equation}
This problem is of course still a \ac{QCQP}.  It is noteworthy, however, that $c$ does not have any constraints on it.  Moreover, the cost is separable  \citep{barham72,kaufman75,golub03} so we can try to optimize for $c$ first and then eliminate it from the problem.  The derivative of the cost with respect to $c$ is
\begin{equation}
\frac{\partial p}{\partial c} = -2 \mbf{b}^T \mbf{n} + 2 c.
\end{equation}
Setting this to zero for an optimum (we should check it is a minimum) we have that
\begin{equation}\label{eq:cfromn}
c = \mbf{b}^T \mbf{n},
\end{equation}
in terms of $\mbf{n}$; we see that this is the projection of the centroid vector onto the line's unit normal vector.  Plugging this back into the cost function we have that
\begin{equation}
p(\mbf{n}) = \mbf{n}^T \mbf{A} \mbf{n} - 2 \mbf{n}^T \mbf{b} \mbf{b}^T \mbf{n} + \mbf{n}^T \mbf{b} \mbf{b}^T \mbf{n} = \mbf{n}^T \underbrace{\left( \mbf{A} - \mbf{b} \mbf{b}^T \right)}_{\mbf{D}} \mbf{n} = \mbf{n}^T \mbf{D} \mbf{n},
\end{equation}
where $\mbf{D}$ is now the covariance of the data with respect to the centroid, $\mbf{b}$; we have essentially used the parallel-axis theorem to transform the covariance from the origin to the centroid.  With $c$ now eliminated, our optimization problem is
\begin{equation}
\begin{array}{rc}
\mbox{min} & p(\mbf{n}) = \mbf{n}^T \mbf{D} \mbf{n} \\
\mbox{w.r.t.} & \mbf{n} \\
\mbox{s.t.} & \mbf{n}^T \mbf{n} = 1
\end{array} .
\end{equation}
Once we solve this problem for $\mbf{n}$, we can use~\eqref{eq:cfromn} to solve for $c$.

\begin{figure}[t]
\centering
\includegraphics[height=0.3\textwidth]{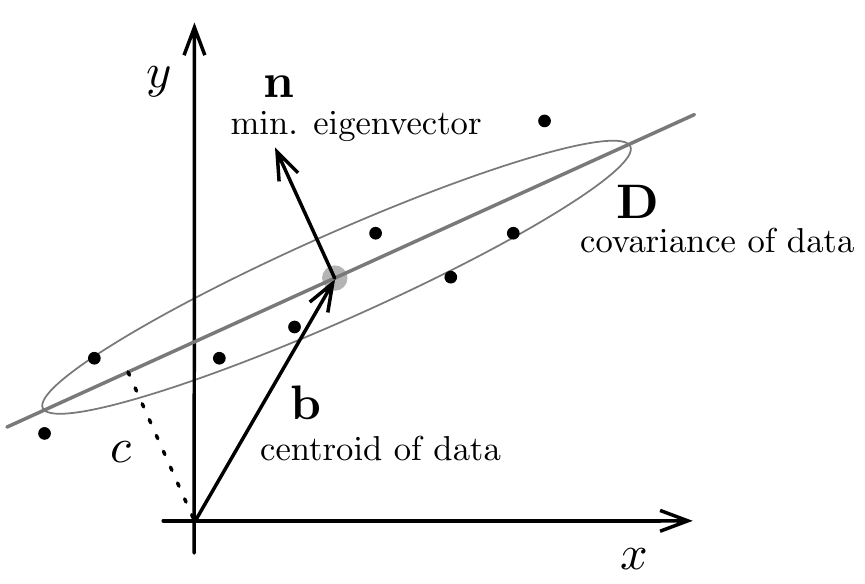}
\caption{Solving \ac{TLS} as an eigenproblem.  The eigenproblem approach finds the centroid, $\mbf{b}$, and covariance, $\mbf{D}$, of the datapoints.  The unit normal, $\mbf{n}$, is an eigenvector in the direction of the smallest eigenvalue of the covariance (direction of least spread).  The offset, $c$, is the projection of the centroid vector, $\mbf{b}$, onto the unit normal.}
\label{fig:eigenproblem}
\end{figure}

The simplest approach to solving~\eqref{eq:eigenproblem} is to use the method of Lagrange to turn our constrained optimization problem into an unconstrained optimization problem.  We can form the {\em Lagrangian} as
\begin{equation}\label{eq:lagrangian}
L(\mbf{n}, \lambda) = \mbf{n}^T \mbf{D} \mbf{n} + \lambda \left( 1 - \mbf{n}^T \mbf{n} \right) = \lambda + \mbf{n}^T \left( \mbf{D} - \lambda \mbf{I} \right) \mbf{n}.
\end{equation}
where $\lambda$ is a {\em Lagrange multiplier}.  Our unconstrained optimization problem is now
\begin{equation}
\begin{array}{rc}
\mbox{min} & L(\mbf{n}, \lambda) = \lambda + \mbf{n}^T \left( \mbf{D} - \lambda \mbf{I} \right) \mbf{n} \\
\mbox{w.r.t.} & \mbf{n}, \lambda \\
\end{array} .
\end{equation}
Taking the derivative of $L$ with respect to $\lambda$ simply gives back our constraint and the derivative with respect to $\mbf{n}$ is
\begin{equation}
\frac{\partial L}{\partial \mbf{n}} =  2 \mbf{n}^T \left( \mbf{D} - \lambda \mbf{I} \right).
\end{equation}
Taking the transpose and setting to zero we have 
\begin{equation}\label{eq:eigenproblem}
\mbf{D} \mbf{n} = \lambda \mbf{n},
\end{equation}
which we can immediately recognize as an {\em eigenproblem}.  Plugging this back into~\eqref{eq:lagrangian}, we have that $L = \lambda $ at an optimum, so to minimize $L$ we want to find the minimum eigenvalue and $\mbf{n}$ will be the associated unit-length eigenvector.  Thus the minimizing cost to our original problem is $p^* = \lambda_{\rm min}(\mbf{D})$, where $ \lambda_{\rm min}(\cdot)$ indicates the minimum eigenvalue of the matrix argument.  Figure~\ref{fig:eigenproblem} depicts the eigenproblem graphically.

\subsection{Lagrangian Dual}

While our {\em primal} optimization problem in~\eqref{eq:eigenproblem} is not {\em convex}, we can form the {\em Lagrangian dual} of our problem \citep{boyd04}, which is guaranteed to be a {\em concave} optimization problem.  The {\em dual function} for this problem is
\begin{equation}
d(\lambda) = \inf_{\mbf{n}} L(\mbf{n}, \lambda) = \left\{  
\begin{array}{cl}
\lambda & \mbox{if $\mbf{D} - \lambda \mbf{I} \geq 0$} \\
-\infty & \mbox{otherwise}
\end{array}
 \right. .
\end{equation}
In this case, the dual problem is fairly easy to solve; the feasible values for $\lambda$ are all those less than or equal to the minimum eigenvalue of $\mbf{D}$.  The optimal cost from the dual problem, $d^*$, is then
\begin{equation}
d^* = \max_\lambda d(\lambda) = \lambda_{\rm min}(\mbf{D}).
\end{equation}
By {\em weak duality}, this is a lower bound on the optimal value of the primal problem, $p^*$:
\begin{equation}
d^* \leq p^*.
\end{equation}
However, we already solved the primal problem in closed form via the eigenproblem in~\eqref{eq:eigenproblem}:
\begin{equation}
\mbf{D} \mbf{n} = \lambda \mbf{n},
\end{equation}
where we argued that we needed to pick $\lambda$ to be the minimum eigenvalue to make $p$ as small as possible.
We therefore have
\begin{equation}
d^* = p^* = \lambda_{\rm min}(\mbf{D}),
\end{equation}
and thus {\em strong duality} holds (i.e., there is no gap between the primary and dual problems).  This property can be used to test for convergence of any solution of the primal problem to the global optimum.

\subsubsection{Dual of the Dual}

We can take things one step further and pose the dual of our dual optimization problem from the previous section.  To do this we form another Lagrangian for the dual problem as
\begin{equation}
L^\prime(\lambda, \mbf{N}) = \lambda + \mbox{tr}\left(\mbf{N}\left(\mbf{D} - \lambda \mbf{I} \right) \right) = \mbox{tr}(\mbf{N} \mbf{D}) + \lambda( 1- \mbox{tr}(\mbf{N})),
\end{equation}
where $\mbf{N} \geq 0$ is a (symmetric since $\mbf{D}$ symmetric) Lagrange multiplier to enforce that $\mbf{D} - \lambda \mbf{I} \geq 0$.  This time we are trying to maximize.  The dual of the dual function is then
\begin{equation}
p^\prime(\mbf{N}) = \sup_{\lambda} L^\prime(\lambda, \mbf{N}) = \left\{ \begin{array}{cl}
 \mbox{tr}(\mbf{N} \mbf{D}) & \mbox{if $\mbox{tr}(\mbf{N})  = 1$} \\
\infty & \mbox{otherwise}
\end{array}
 \right. .
\end{equation}
We assume the first condition so that we have the following optimization problem:
\begin{equation}\label{eq:sdp}
\begin{array}{rc}
\mbox{min} & p^\prime(\mbf{N}) = \mbox{tr}(\mbf{N} \mbf{D}) \\
\mbox{w.r.t.} & \mbf{N} \\
\mbox{s.t.} & \mbox{tr}(\mbf{N}) = 1 \\
& \mbf{N} \geq 0 
\end{array} .
\end{equation}
This is actually a {\em relaxation} of our problem in~\eqref{eq:eigenproblem}.  To see this, we note that 
\begin{equation}
p(\mbf{n}) = \mbf{n}^T \mbf{D} \mbf{n} = \mbox{tr}(\mbf{n} \mbf{n}^T \mbf{D}),
\end{equation}
and
\begin{equation}
 \mbox{tr}(\mbf{n}\mbf{n}^T) = \mbf{n}^T \mbf{n} = 1.
\end{equation}
Thus if we let $\mbf{N} = \mbf{n}\mbf{n}^T$ then our primal problem can be written exactly as
\begin{equation}\label{eq:primal}
\begin{array}{rc}
\mbox{min} & p(\mbf{N}) = \mbox{tr}(\mbf{N} \mbf{D}) \\
\mbox{w.r.t.} & \mbf{N} \\
\mbox{s.t.} & \mbox{tr}(\mbf{N})  = 1 \\
& \mbox{rank}(\mbf{N}) = 1 \\
& \mbf{N} \geq 0 
\end{array} .
\end{equation}
If we drop the rank constraint, we have the problem in~\eqref{eq:sdp}.  It is a relaxation in the sense that the feasible values for $\mbf{N}$ contain all the feasible values for $\mbf{n}\mbf{n}^T$ in our original problem.  Therefore, the optimal solution to our original problem is contained in the relaxation.  The relaxed problem is called a {\em \acf{SDP}} and it is convex; it has a linear objective and linear constraints.  

Thus, rather than solving the eigenproblem of~\eqref{eq:eigenproblem}, we can try solving the \ac{SDP} of~\eqref{eq:sdp}.  If the solution has $\mbox{rank}(\mbf{N}) = 1$, then we have solved the original problem exactly; otherwise we can use a rank-$1$ truncated {\em \ac{SVD}} to enforce the correct rank and extract $\mbf{n}$ from $\mbf{N}$. 

\subsubsection{Geometric Interpretation}

\begin{figure}[t]
\centering
\includegraphics[height=0.25\textwidth]{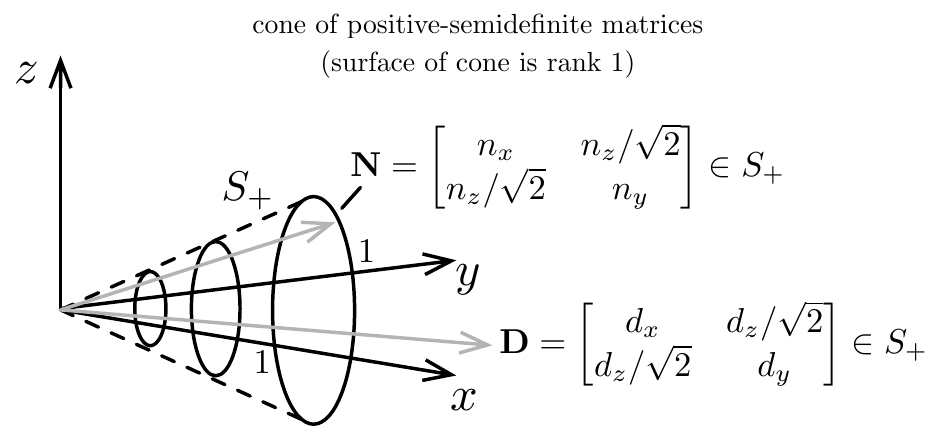}
\caption{Visualization of the cone of $2 \times 2$ \ac{PSD} matrices.  The axial angle is $45$ degrees.  All points inside this cone (a convex set) represent \ac{PSD} matrices.}
\label{fig:cone}
\end{figure}

To understand why the relaxation (without the rank constraint) in~\eqref{eq:sdp} works, we can look at the problem geometrically.  Our two matrices, $\mbf{N}$ and $\mbf{D}$, are both \ac{PSD}.  Figure~\ref{fig:cone} shows that all $2 \times 2$ \ac{PSD} matrices form a cone, $S_+$, with axial angle of $45$ degrees in three-dimensional space.  Our data matrix, $\mbf{D}$, is a \ac{PSD} covariance matrix and thus lives inside this cone.

Our constraint, $\mbox{tr}(\mbf{N}) = 1$, represents a plane in Figure~\ref{fig:cone} with equation $x + y = 1$.  The intersection of $S_+$ and this plane is a circular disk and $\mbf{N}$ must lie in this disk (a convex set).  Using the parameterization of $\mbf{N}$ and $\mbf{D}$ in the figure, our objective in~\eqref{eq:sdp} is to minimize
\begin{equation}
	\mbox{tr}(\mbf{N} \mbf{D}) = \bbm n_x \\ n_z/\sqrt{2} \\ n_z/\sqrt{2} \\ n_y \ebm^T \bbm d_x \\ d_z/\sqrt{2} \\ d_z/\sqrt{2} \\ d_y \ebm  = \bbm n_x \\ n_y \\ n_z \ebm^T \bbm d_x \\ d_y \\ d_z \ebm .
\end{equation}
We see this is just the inner product of two vectors in three-dimensional space.  Since we know that the tip of $\mbf{N}$ (as a vector) lies in the circular disk already mentioned, our objective is to make the angle between $\mbf{N}$ and $\mbf{D}$ as large as possible.  The unique solution is to put $\mbf{N}$ at the far side of the disk, as far away as possible from $\mbf{D}$.  This has the effect of putting $\mbf{N}$ on the surface of the $S_+$ cone, and therefore it will automatically have rank $1$.  This explains why our dropping the rank condition in~\eqref{eq:sdp} works; the relaxed problem is {\em tight} meaning it will always find a rank $1$ solution.  We can attempt to use this same concept in other problems such as the robust line-fitting problem that we treat next.

\section{Robust Line Fitting}

The previous section on \ac{TLS} assumed that all the datapoints were reasonable.  In practice, there may be outlier datapoints that do not conform well to the line that we are trying to fit.  To handle outliers, we can turn to {\em M-estimation} (see e.g., \citet[\S 5.3.2]{barfoot_ser17}) and use a {\em robust cost function} \citep{zhang97,mactavish_crv15}.

In M-estimation, we replace the overall objective that we seek to minimize with
\begin{equation}\label{eq:robust}
p(a,b,c) = \sum_n \rho( \underbrace{ax_n + b y_n - c}_{e_n}),
\end{equation}
where $\rho( e )$ is a robust cost function.  For the purpose of this document, we will pick the {\em \ac{GM}} robust cost,
\begin{equation}
\rho(e) = \frac{e^2}{1 + e^2}.
\end{equation}
We see that the larger $e^2$ is, the closer the $\rho(e)$ is to $1$, effectively down-weighting outlier measurements in the overall cost function.

This is where things get interesting from an optimization point of view.  While the regular \ac{TLS} problem was not convex, it was still relatively easy to solve it, for example using the eigenproblem approach; this is because there is only a single minimum.  In the robust version, our problem can have local minima and we would like to be able to find the global minimum amongst these, or at least know whether a solution is a local or global minimum. 

The next section will introduce {\em \acf{IRLS}} as the typical way of solving our M-estimation problem; this approach can converge to a local minimum.  Following that, we show how to pose the problem as an \ac{SDP} that can be solved globally using standard solvers, but these can become slow with a lot of datapoints.  Finally, we attempt to get the best of both worlds by showing how we can certify the \ac{IRLS} solution as global in most cases.

\subsection{Iteratively Reweighted Least-Squares}

In this section, we show the standard approach to M-estimation, which is to use \ac{IRLS} \citep{holland77}.  The idea is replace~\eqref{eq:robust} with a weighted version,
\begin{equation}\label{eq:irls}
p(a,b,c) = \sum_n w_n \, ( ax_n + b y_n - c)^2,
\end{equation}
where $w_n$ is a weight for each datapoint that will be iteratively updated.  To ensure that~\eqref{eq:robust} and~\eqref{eq:irls} have the same minima, we want the first-order optimality conditions to be the same.  This requires us to select
\begin{equation}\label{eq:weight}
w_n = \frac{1}{e_n} \left. \frac{\partial \rho}{\partial e} \right|_{e = e_n} = \frac{2}{(1+ e_n^2)^2},
\end{equation}
for the \ac{GM} robust cost function.  Since we are iterating, we set the $w_n$ based on the values of $e_n$ from the previous iteration and then hold them fixed.  This makes our problem look very similar to~\eqref{eq:tls}. 

With the weights fixed our objective is
\begin{equation}
p(a,b,c) = \underbrace{\bbm a \\ b \ebm^T}_{\mbf{n}^T} \underbrace{\bbm \sum_n w_n x_n^2 & \sum_n w_n x_n y_n \\ \sum_n w_n x_n y_n &  \sum_n w_n y_n^2 \ebm}_{\mbf{A}} \underbrace{\bbm a \\ b \ebm}_{\mbf{n}} - 2 c \underbrace{\bbm  \sum_n w_n x_n \\  \sum_n w_n y_n \ebm^T}_{\mbf{b}^T}\underbrace{\bbm a \\ b \ebm}_{\mbf{n}} +  \underbrace{\sum_n w_n}_{w} \, c^2.
\end{equation}
We can again eliminate $c$ from the problem by optimizing it first:
\begin{equation}
\frac{\partial p}{\partial c} = -2 \mbf{b}^T \mbf{n} + 2 w c.
\end{equation}
Setting this to zero we have
\begin{equation}\label{eq:cfromn2}
c = \frac{1}{w} \mbf{b}^T \mbf{n},
\end{equation}
in terms of $\mbf{n}$.  Substituting this back into $p$ to eliminate $c$ we have
\begin{equation}
p(\mbf{n}) = \mbf{n}^T \underbrace{\left( \mbf{A} - \frac{1}{w} \mbf{b} \mbf{b}^T \right)}_{\mbf{D}} \mbf{n} = \mbf{n}^T \mbf{D} \mbf{n}.
\end{equation}
The optimization problem for the unit normal is then
\begin{equation}
\begin{array}{rc}
\mbox{min} & p(\mbf{n}) = \mbf{n}^T \mbf{D} \mbf{n} \\
\mbox{w.r.t.} & \mbf{n} \\
\mbox{s.t.} & \mbf{n}^T \mbf{n} = 1
\end{array} ,
\end{equation}
which we can solve first then use~\eqref{eq:cfromn2} to find $c$.  This process must be iterated now with the weights recalculated from~\eqref{eq:weight} each time.  We can initialize at the first iteration by setting $w_n = 1$, which will result in the same solution as the regular \ac{TLS} problem from the previous section.  However, this initialization can result in the algorithm converging to a poor local minimum.

\subsection{Semi-Definite Program}

If we want to ensure that we are finding the global minimum of~\eqref{eq:irls}, we need to try a different approach than \ac{IRLS}.  One possibility is to use a similar approach to~\citet{yang20}.  Our optimization problem, with the \ac{GM} robust cost function, is
\begin{equation}\label{eq:robusttls}
\begin{array}{rc}
\mbox{min} & p(a,b,c) = \sum_n \frac{( ax_n + b y_n - c)^2}{1 + ( ax_n + b y_n - c)^2} \\
\mbox{w.r.t.} & a,b,c \\
\mbox{s.t.} & a^2 + b^2 = 1
\end{array} .
\end{equation}
We can use the device of \citet{black96} to restate this optimization problem as
\begin{equation}\label{eq:black}
\begin{array}{rc}
\mbox{min} & p(a,b,c,\alpha_n) =  \sum_n \alpha_n^2 ( ax_n + b y_n - c)^2 + (\alpha_n - 1)^2 \\
\mbox{w.r.t.} & a,b,c,\alpha_n \\
\mbox{s.t.} & a^2 + b^2 = 1
\end{array} .
\end{equation}
If we optimize this problem for the $\alpha_n$ first to eliminate them, we recover~\eqref{eq:robusttls}.  Taking the derivative of $p(a,b,c,\alpha_n)$ with respect to $\alpha_n$ we have
\begin{equation}
\frac{\partial p}{\partial \alpha_n} = 2 \alpha_n e_n^2 + 2 (\alpha_n - 1).
\end{equation}
Setting this to zero and isolating for $\alpha_n$ gives
\begin{equation}
\alpha_n = \frac{1}{1 + e_n^2},
\end{equation}
which we note must satisfy $0 < \alpha_n \leq 1$.  Plugging this back into~\eqref{eq:black} we arrive at~\eqref{eq:robusttls}.  This problem is not a \ac{QCQP}, but we can manipulate it into one.  

First, we will define some new variables:
\begin{equation}
\mbf{q}_n = \alpha_n \mbf{q}_0 = \alpha_n \bbm a \\ b \\ c \ebm , \quad \mbf{d}_n = \bbm x_n \\ y_n \\ -1 \ebm,
\end{equation}
where $n=1 \ldots N$.
Since we want $\mbf{n}^T \mbf{n} = 1$, we have the useful results,
\begin{equation}
\mbf{q}_n^T \mbf{J} \mbf{q}_n = \alpha_n^2, \quad \mbf{q}_n^T \mbf{J} \mbf{q}_0 = \alpha_n, \quad \mbf{q}_0^T \mbf{J} \mbf{q}_0 = 1,
\end{equation}
where
\begin{equation}
\mbf{J} = \bbm 1 & 0 & 0 \\ 0 & 1 & 0 \\ 0 & 0 & 0 \ebm .
\end{equation}
Our objective can then be written as
\begin{equation}
p(\mbf{q}_0, \mbf{q}_1, \ldots, \mbf{q}_n) = \sum_n \mbf{q}_n^T \left( \mbf{J} + \mbf{d}_n \mbf{d}_n^T \right) \mbf{q}_n - 2 \mbf{q}_n^T \mbf{J} \mbf{q}_0 + \mbf{q}_0^T \mbf{J} \mbf{q}_0 = \mbf{q}^T \mbf{H} \mbf{q},
\end{equation}
where
\begin{equation}
\mbf{q} = \bbm \mbf{q}_0 \\ \mbf{q}_1 \\ \mbf{q}_2 \\ \vdots \\ \mbf{q}_N \ebm , \quad 
\mbf{H} = \bbm N \mbf{J} & -\mbf{J} & -\mbf{J} & \cdots & -\mbf{J} \\ -\mbf{J} & \mbf{J} + \mbf{d}_1 \mbf{d}_1^T & \mbf{0} & \cdots & \mbf{0} \\ -\mbf{J} & \mbf{0} & \mbf{J} + \mbf{d}_2 \mbf{d}_2^T & \cdots & \mbf{0} \\ \vdots & \vdots & \vdots & \ddots & \vdots \\ -\mbf{J} & \mbf{0} & \mbf{0} & \cdots & \mbf{J} + \mbf{d}_N \mbf{d}_N^T \ebm.
\end{equation}
We note that $\mbf{H}$ is an arrowhead matrix in passing.

With this version of our cost function, we can write the optimization problem as
\begin{equation}\label{eq:robustqcqp}
\begin{array}{rc}
\mbox{min} & p(\mbf{q}) =  \mbf{q}^T \mbf{H} \mbf{q} \\
\mbox{w.r.t.} & \mbf{q} \\
\mbox{s.t.} & \mbf{q}_0^T \mbf{J} \mbf{q}_0 = 1 \\
& (\forall n) \; \mbf{q}_0 \mbf{q}_n^T = \mbf{q}_n \mbf{q}_0^T 
\end{array} .
\end{equation}
This is now a \ac{QCQP}, with several equality constraints.  The first equality constraint simply ensures our original $a^2 + b^2 = 1$ unit-normal requirement.  The other constraints ensure that all the $\mbf{q}_n$ are parallel to $\mbf{q}_0$ as required by $\mbf{q}_n = \alpha_n \mbf{q}_0$.  There is one other small issue to resolve, which is that the third column (and row) of $\mbf{H}$ is all zeros; a workaround to this is to install a small number such as $10^{-6}$ in the $(3,3)$ entry of $\mbf{H}$, which amounts to putting a prior on $c$ to keep the line from being extremely far from the origin.

With several equality constraints, solving~\eqref{eq:robustqcqp} directly is difficult.  Instead, we will again relax our problem by noting that $\mbf{q}^T \mbf{H} \mbf{q} = \mbox{tr}(\mbf{q}\mbf{q}^T \mbf{H} )$.  Then if we let $\mbf{Q}$ replace $\mbf{q}\mbf{q}^T$, a relaxed version of our problem is
\begin{equation}\label{eq:robustsdp}
\begin{array}{rc}
\mbox{min} & p(\mbf{Q}) =  \mbox{tr}( \mbf{Q}\mbf{H} ) \\
\mbox{w.r.t.} & \mbf{Q} \\
\mbox{s.t.} & \mbox{tr}( \mbf{Q}_{00} \mbf{J} ) = 1 \\
& (\forall n) \; \mbf{Q}_{n0} = \mbf{Q}_{n0}^T \\
& \mbf{Q} \geq 0 
\end{array} ,
\end{equation}
which is a \ac{SDP}.  With $\mbf{Q}$ broken into $3 \times 3$ blocks, the notation $\mbf{Q}_{ij}$ refers to the block in the $ij$ location.  Unfortunately, this \ac{SDP} is now too relaxed and not well posed.  We can introduce some additional {\em redundant constraints} \citep{yang20} that do not affect the original problem but make the relaxed \ac{SDP} situation improve.  A functional set of constraints is captured in this version of the problem:
\begin{equation}\label{eq:robustsdp2}
\begin{array}{rc}
\mbox{min} & p(\mbf{Q}) =  \mbox{tr}( \mbf{Q}\mbf{H} ) \\
\mbox{w.r.t.} & \mbf{Q} \\
\mbox{s.t.} & \mbox{tr}( \mbf{Q}_{00} \mbf{J} ) = 1 \\
& (\forall n, m < n) \; \mbf{Q}_{nm} = \mbf{Q}_{nm}^T \\
& \mbf{Q} \geq 0 
\end{array} ,
\end{equation}
The second constraint has now been expanded to ensure that all the $\mbf{q}_n$ are parallel to each other everywhere they are represented in $\mbf{Q}$.  
This problem can now be solved using any standard \ac{SDP} solver, which can find the global minimum fairly easily.  If the minimizing $\mbf{Q}$ (to problem~\eqref{eq:robustsdp2}) is rank $1$, then we have also found the global minimum of~\eqref{eq:robustqcqp} since this implies $\mbf{Q} = \mbf{q}\mbf{q}^T$, which we can extract using a \ac{SVD}.  Unfortunately, as the number of datapoints, $N$, becomes large (e.g., hundreds), problem~\eqref{eq:robustsdp2} becomes too expensive to solve with a general \ac{SDP} solver.

\subsection{Optimality Certificate}

We would like to have the best of both worlds:  the speed of \ac{IRLS} and the global optimality guarantee of the \ac{SDP} problem.  A compromise might be to solve the problem using \ac{IRLS} (fast and usually finds the global minimum) and then test if we are at the global minimum some other way.  We refer to this as a {\em certificate of optimality}.   To find a certificate we can again turn to Lagrangian duality starting from~\eqref{eq:robustqcqp} with the redundant constraints of~\eqref{eq:robustsdp2} added in.  We will closely follow the approach of \citet{yang20} in this section.

An equivalent condition to the constraint, $\mbf{q}_n \mbf{q}_m^T = \mbf{q}_m \mbf{q}_n^T$, is that $\mbf{q}_n$ and $\mbf{q}_m$ are parallel or $\mbf{q}_n^\times \mbf{q}_m = \mbf{0}$ (cross product is the zero vector).  Our \ac{QCQP} with redundant constraints can then be stated as
\begin{equation}\label{eq:robustqcqp2}
\begin{array}{rc}
\mbox{min} & p(\mbf{q}) =  \mbf{q}^T \mbf{H} \mbf{q} \\
\mbox{w.r.t.} & \mbf{q} \\
\mbox{s.t.} & \mbf{q}_0^T \mbf{J} \mbf{q}_0 = 1 \\
& (\forall n, m < n) \; \mbf{q}_n^\times \mbf{q}_m = \mbf{0} 
\end{array} .
\end{equation}
This problem should have the same dual as~\eqref{eq:robustsdp2}.

We can form the Lagrangian for this problem as
\begin{equation}
L(\mbf{q}, \lambda, \mbs{\gamma}_1, \ldots, \mbs{\gamma}_N) = \mbf{q}^T \mbf{H} \mbf{q} + \lambda(1 - \mbf{q}_0^T \mbf{J} \mbf{q}_0) + 2 \sum_{n, m < n}  \mbs{\gamma}_{nm}^T \mbf{q}_n^\times \mbf{q}_m = \lambda + \mbf{q}^T \mbf{K} \mbf{q},
\end{equation}
where $\lambda, \mbs{\gamma}_{nm}$ are Lagrange multipliers for our $N(N+1)/2 + 1$ equality constraints and
\begin{equation} \label{eq:K}
\mbf{K} = \bbm (N - \lambda) \mbf{J} & -\mbf{J} - \mbs{\gamma}_{10}^{\times^T} & -\mbf{J}- \mbs{\gamma}_{20}^{\times^T}  & \cdots & -\mbf{J} - \mbs{\gamma}_{N0}^{\times^T}  \\ -\mbf{J} - \mbs{\gamma}_{10}^{\times}  & \mbf{J} + \mbf{d}_1 \mbf{d}_1^T & -\mbs{\gamma}_{21}^{\times^T} & \cdots & -\mbs{\gamma}_{N1}^{\times^T} \\ -\mbf{J} - \mbs{\gamma}_{20}^{\times}& -\mbs{\gamma}_{21}^{\times} & \mbf{J} + \mbf{d}_2 \mbf{d}_2^T & \cdots & -\mbs{\gamma}_{N2}^{\times^T} \\ \vdots & \vdots & \vdots & \ddots & \vdots \\ -\mbf{J} - \mbs{\gamma}_{N0}^{\times} & -\mbs{\gamma}_{N1}^{\times} & -\mbs{\gamma}_{N2}^{\times} & \cdots & \mbf{J} + \mbf{d}_N \mbf{d}_N^T \ebm,
\end{equation}
which is no longer arrowhead and depends on the Lagrange multipliers.

The dual function is
\begin{equation}
d(\lambda) = \inf_{\mbf{n}} L(\mbf{n}, \lambda) = \left\{  
\begin{array}{cl}
\lambda & \mbox{if $\mbf{K} \geq 0$} \\
-\infty & \mbox{otherwise}
\end{array}
 \right. .
\end{equation}

The first-order optimality condition for this problem is then simply
\begin{equation}\label{eq:robustoptcond}
\frac{\partial p}{\partial \mbf{q}^T} = \mbf{K} \mbf{q} = \mbf{0}.
\end{equation}
Suppose now that we have a candidate solution for $\mbf{q}$ and we wish to try to certify the solution.  If the solution is globally optimal, we should be able to find a $\mbf{K}$ that has the structure in~\eqref{eq:K}, satisfies the first-order optimality conditions $\mbf{K} \mbf{q} = \mbf{0}$, and is positive-semidefinite, $\mbf{K} \geq 0$.  Since $\lambda$ represents the final cost of the candidate solution, we can hold this multiplier fixed and search for $\mbs{\gamma}_{nm}$ to try to satisfy the required conditions.  If we can do so, then our solution is globally optimal.  If we cannot, we cannot certify the solution; in other words, the test is sufficient but not necessary.

It turns out to be difficult to solve for a set of $\mbs{\gamma}_{nm}$ directly.  Instead, we can use a technique called {\em Douglas-Rachford splitting} \citep{douglas56, lions79}, which roughly says that if you want to find something in the (nonempty) intersection of two convex sets, you can alternately project a candidate onto one set then the other until you are in both sets.  

The $\mbf{K} \geq 0$ condition defines our first convex set, the positive-semidefinite cone of matrices, $S_+$.  Given a $\mbf{K}$ real and symmetric but not in $S_+$, we can project it there by first carrying out a spectral decomposition (i.e., eigendecomposition),
\begin{equation}
\mbf{K} = \sum_n \lambda_n \mbf{v}_n \mbf{v}_n^T,
\end{equation}
where $\lambda_n$ are the (real) eigenvalues and $\mbf{v}_n$ the (orthonormal) eigenvectors.   Then, the closest matrix to $\mbf{K}$ that is in $S_+$ keeps only those modes with nonnegative eigenvalues.  Our projection then becomes
\begin{equation}
\mbf{K}^\prime =  \mbox{proj}_{S_+} \mbf{K} =  \sum_n \mbox{max}(0,\lambda_n) \mbf{v}_n \mbf{v}_n^T,
\end{equation}
which is relatively easy to compute.

The first-order optimality conditions, $\mbf{K}\mbf{q} = \mbf{0}$ and the required structure of $\mbf{K}$ in~\eqref{eq:K} together form a set of linear constraints that define a subspace (a convex set) of all possible matrices, $\mbf{K}$.  We need to work out how to project a candidate $\mbf{K}$ to this subspace.  We can rearrange the first-order optimality conditions as
\begin{equation}
\underbrace{\bbm \mbf{0} &  \mbs{\gamma}_{10}^{\times^T} & \mbs{\gamma}_{20}^{\times^T}  & \cdots &  \mbs{\gamma}_{N0}^{\times^T}  \\ \mbs{\gamma}_{10}^{\times}  & \mbf{0} & \mbs{\gamma}_{21}^{\times^T} & \cdots & \mbs{\gamma}_{N1}^{\times^T} \\  \mbs{\gamma}_{20}^{\times}& \mbs{\gamma}_{21}^{\times} & \mbf{0} & \cdots & \mbs{\gamma}_{N2}^{\times^T} \\ \vdots & \vdots & \vdots & \ddots & \vdots \\ \mbs{\gamma}_{N0}^{\times} & \mbs{\gamma}_{N1}^{\times} & \mbs{\gamma}_{N2}^{\times} & \cdots & \mbf{0} \ebm}_{\mbs{\Gamma}}   \bbm \mbf{q}_0 \\ \mbf{q}_1 \\ \mbf{q}_2 \\ \vdots \\ \mbf{q}_N \ebm = \underbrace{\bbm (N-\lambda) \mbf{J} & -\mbf{J} & -\mbf{J} & \cdots & -\mbf{J} \\ -\mbf{J} & \mbf{J} + \mbf{d}_1 \mbf{d}_1^T & \mbf{0} & \cdots & \mbf{0} \\ -\mbf{J} & \mbf{0} & \mbf{J} + \mbf{d}_2 \mbf{d}_2^T & \cdots & \mbf{0} \\ \vdots & \vdots & \vdots & \ddots & \vdots \\ -\mbf{J} & \mbf{0} & \mbf{0} & \cdots & \mbf{J} + \mbf{d}_N \mbf{d}_N^T \ebm}_{\mbf{M}} \bbm \mbf{q}_0 \\ \mbf{q}_1 \\ \mbf{q}_2 \\ \vdots \\ \mbf{q}_N \ebm,
\end{equation}
where everything on the right is now known.  Clearly there is not enough information here to solve for the $\mbs{\gamma}_{nm}$ uniquely since there are $N+1$ block equations and $N(N+1)/2$ unknown Lagrange multipliers.  However, owing to the cross-product property, $\mbf{u}^\times \mbf{v} = -\mbf{v}^\times \mbf{u}$, we can rewrite the left-hand side as
\begin{equation}
\mbs{\Gamma}\mbf{q} = \mbf{R} \mbs{\gamma},
\end{equation}
where $\mbs{\gamma}$ stacks all the vector versions of the Lagrange multipliers and $\mbf{R}$ depends only on $\mbf{q}$.  For example, with $N=3$ we have
\begin{equation}
\mbs{\gamma} = \bbm \mbs{\gamma}_{10} \\ \mbs{\gamma}_{20} \\  \mbs{\gamma}_{30} \\ \mbs{\gamma}_{21} \\ \mbs{\gamma}_{31} \\ \mbs{\gamma}_{32} \ebm, \quad \mbf{R} = \bbm \mbf{q}_1^\times & \mbf{q}_2^\times & \mbf{q}_3^\times & \mbf{0} & \mbf{0} & \mbf{0} \\ -\mbf{q}_0^\times & \mbf{0} & \mbf{0} & \mbf{q}_2^\times & \mbf{q}_3^\times & \mbf{0} \\ \mbf{0} & -\mbf{q}_0^\times & \mbf{0} & -\mbf{q}_1^\times & \mbf{0} & \mbf{q}_3^\times \\ \mbf{0} & \mbf{0} & -\mbf{q}_0^\times & \mbf{0} & -\mbf{q}_1^\times  & -\mbf{q}_2^\times \ebm,
\end{equation}
which should be enough to see the pattern.  This means our constraint on the Lagrange multipliers is the linear system,
\begin{equation}
\mbf{R} \mbs{\gamma}  = \mbf{M} \mbf{q}.
\end{equation}
The steps for projecting a candidate $\mbf{K}$ onto the desired subspace are therefore
\begin{enumerate}
\item compute:  $\mbs{\Gamma} = \mbf{M} - \mbf{K}$,
\item extract:  $\mbs{\gamma}$ from $\mbs{\Gamma}$,
\item project:  $\mbs{\gamma}^\prime = \mbs{\gamma} - \mbf{R}^T \left( \mbf{R} \mbf{R}^T \right)^+ \left( \mbf{R} \mbs{\gamma} - \mbf{M} \mbf{q} \right)$,
\item build: $\mbs{\Gamma}^\prime$ from $\mbs{\gamma}^\prime$,
\item compute:  $\mbf{K}^\prime = \mbf{M} - \mbs{\Gamma}^\prime$.
\end{enumerate}
We will refer to this procedure as $\mbf{K}^\prime = \mbox{proj}_{\rm sub} \mbf{K}$.  Note, the $^+$ indicates the pseudoinverse.

The Douglas-Rachford splitting procedure is then
\beqn{}
\mbf{K}_1 & = & \mbox{proj}_{S_+} \mbf{K}^{(i)}, \\
\mbf{K}_2 & = & \mbox{proj}_{\rm sub} \left( 2 \mbf{K}_1 - \mbf{K}^{(i)} \right), \\
\mbf{K}^{(i+1)} & = & \mbf{K}^{(i)} + \beta \left( \mbf{K}_2 - \mbf{K}_1 \right),
\eeqn
which is stable for $0 < \beta < 2$ and $i$ is an iteration index.  We initialize $\mbs{\Gamma} = \mbf{0}$, although better initial guesses may be possible.  If the intersection of the two convex sets is nonempty, the procedure converges to a point in the intersection.  

If the intersection is empty, then the procedure will not converge.  Thus, if we can achieve convergence, we have satisfied the required conditions for our solution to be certified as globally optimal; this can be checked by looking at the minimum eigenvalue of $\mbf{K}_2$ (right after the projection to the subspace); if it is nonnegative (or suitably close) then we are done and the solution can be certified.

\section{Numerical Example}

\begin{figure}[t]
\centering
\includegraphics[height=0.4\textwidth]{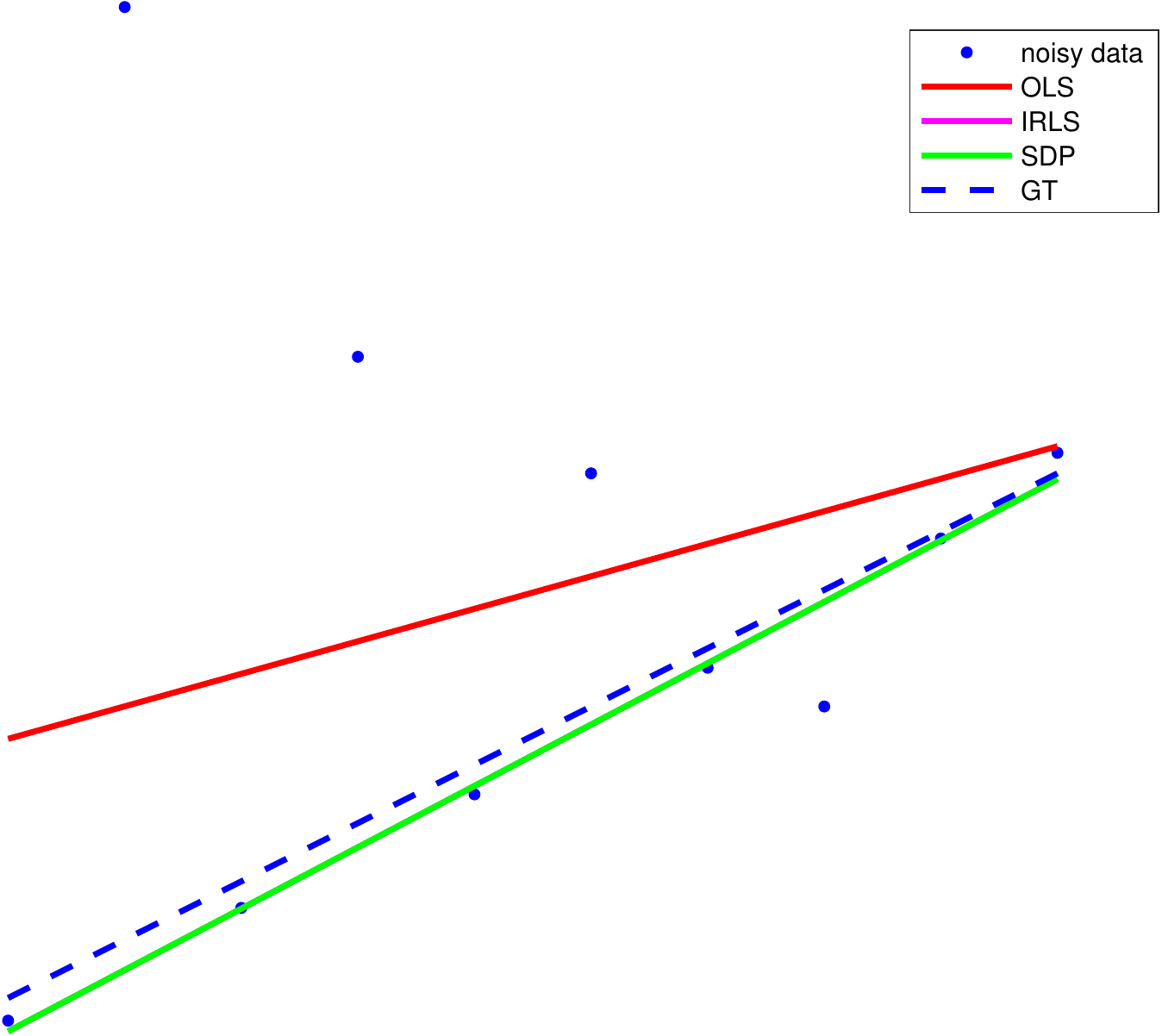}\hspace*{0.5in}
\includegraphics[height=0.4\textwidth]{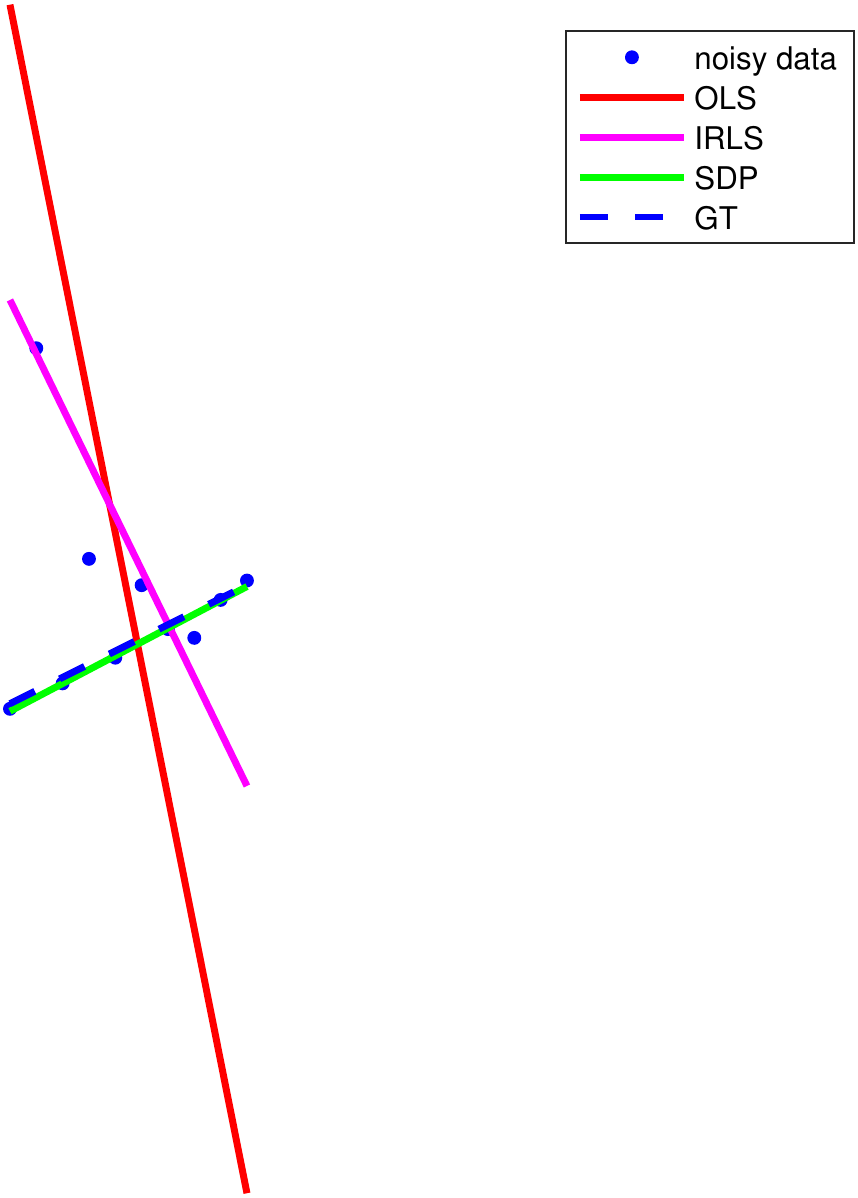}
\caption{Examples of \ac{TLS} line fitting.  There are $N=10$ datapoints with $4$ obvious outliers.  The blue dashed line is the ground-truth for the inlier measurements.  The red line shows what the non-robust line fit does.  Purple (possibly under green) is the \ac{IRLS} solution.  Green is the \ac{SDP} solution, which is the global minimum (solved using CVX with Mosek).  The left example shows \ac{IRLS} converging to the global minimum, while the right example shows it getting stuck in a local minimum.}
\label{fig:tls_examples}
\end{figure}

\begin{figure}[t]
\centering
\includegraphics[height=0.4\textwidth]{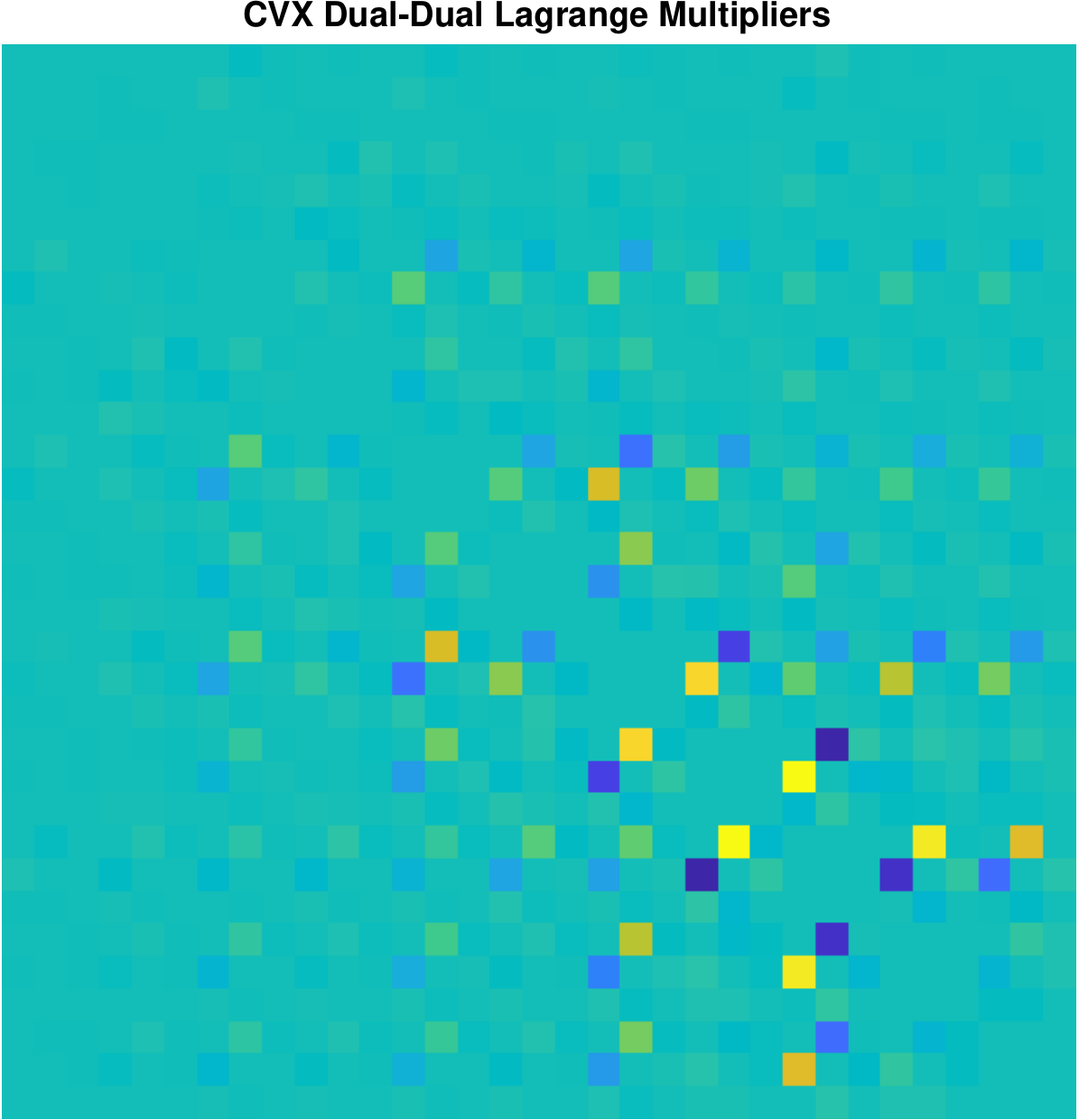}\hspace*{0.5in}
\includegraphics[height=0.4\textwidth]{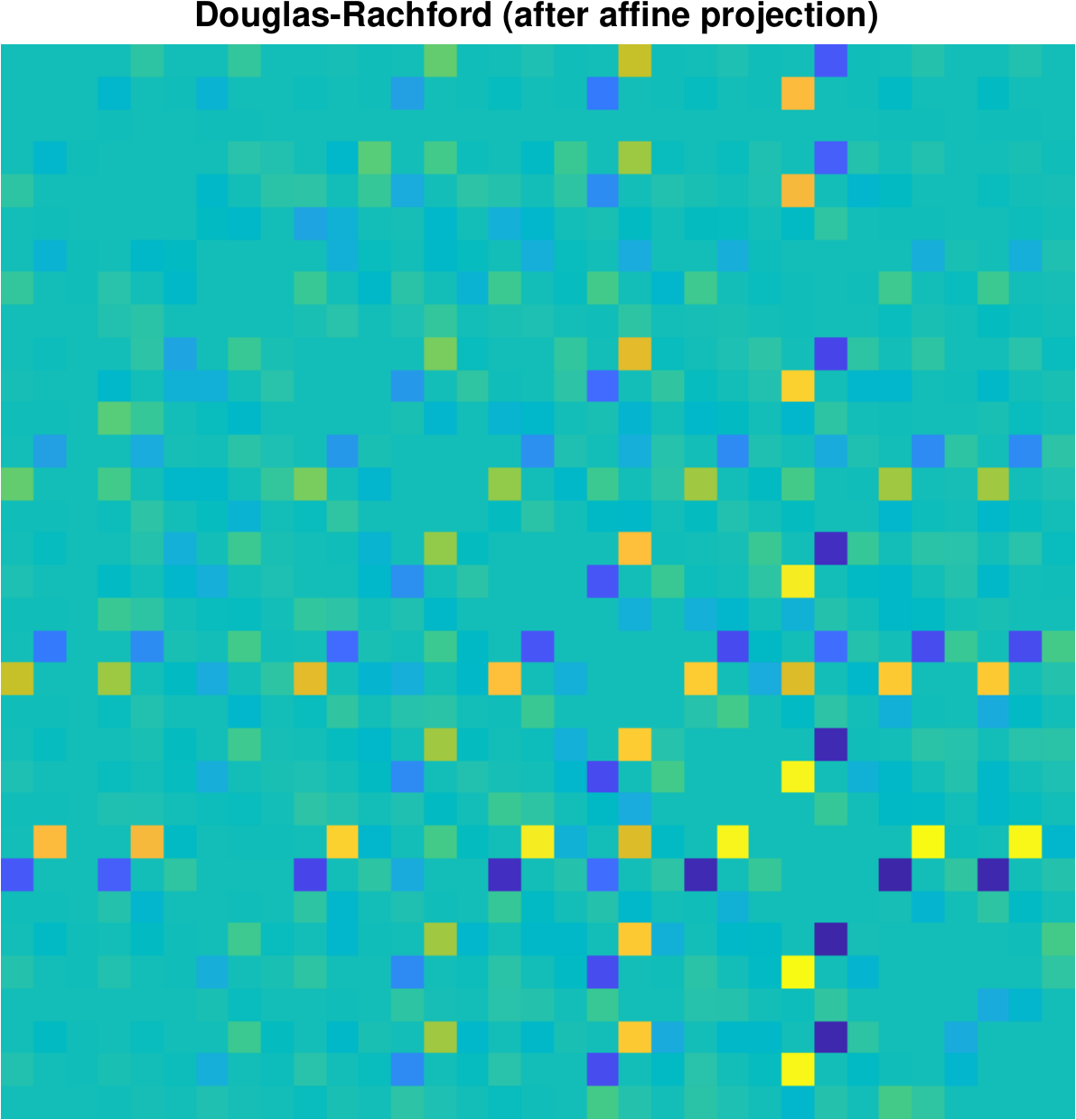}
\caption{Visualization of the $\mbs{\Gamma}$ matrix as computed by (left) the \ac{SDP} solver (CVX with Mosek) and (right) the Douglas-Rachford splitting algorithm for the case where \ac{IRLS} converges to the correct global minimum.  We see they are visually similar but not identical; this is likely because the multipliers are not unique due to the constraints being redundant.}
\label{fig:multipliers}
\end{figure}

Figure~\ref{fig:tls_examples} shows two cases of robust line fitting.  \ac{IRLS} converges to the correct global minimum in one example but a suboptimal one in the other; the Douglas-Rachford splitting procedure is able to certify \ac{IRLS} when it does converge to the correct global minimum (minimum eigenvalue of $\mbf{K}$ is $3.05 \times 10^{-7}$ after $300$ Douglas-Rachford iterations), while it fails to certify in the local minimum case (minimum eigenvalue of $\mbf{K}$ is $-0.42$ after $300$ Douglas-Rachford iterations). Figure~\ref{fig:multipliers} provides a visualization of the $\mbs{\Gamma}$ matrix.

\section{Conclusion}

We have discussed the \ac{TLS} line-fitting problem from a number of different optimization perspectives and shown some modern optimization techniques in action on this old problem.  The main purpose of this technical note is to be tutorial in nature and we welcome any comments to make it more clear and useful.

\input{fineline.bbl}

\end{document}

%% file: notation_header.tex
\newcommand{\bbm}{\begin{bmatrix}}
\newcommand{\ebm}{\end{bmatrix}}
\newcommand{\mbf}{\mathbf}
\newcommand{\mbs}[1]{{\boldsymbol{#1}}}

\newcommand{\beq}{\begin{equation}}
\newcommand{\eeq}{\end{equation}}
\newcommand{\bdis}{\begin{displaymath}}
\newcommand{\edis}{\end{displaymath}}
\newcommand{\beqn}[1]{\begin{subequations}\label{eq:#1}\begin{eqnarray}}
\newcommand{\eeqn}{\end{eqnarray}\end{subequations}}

%% file: fineline.bbl
\begin{thebibliography}{12}
\newcommand{\enquote}[1]{``#1''}
\providecommand{\natexlab}[1]{#1}

\bibitem[{Barfoot(2017)}]{barfoot_ser17}
Barfoot, T.~D., \emph{State Estimation for Robotics}, Cambridge University
  Press, 2017.

\bibitem[{Barham and Drane(1972)}]{barham72}
Barham, R.~H. and Drane, W., \enquote{An algorithm for least squares estimation
  of nonlinear parameters when some of the parameters are linear,}
  \emph{Technometrics}, 14(3):757--766, 1972.

\bibitem[{Black and Rangarajan(1996)}]{black96}
Black, M.~J. and Rangarajan, A., \enquote{On the unification of line processes,
  outlier rejection, and robust statistics with applications in early vision,}
  \emph{International journal of computer vision}, 19(1):57--91, 1996.

\bibitem[{Boyd and Vandenberghe(2004)}]{boyd04}
Boyd, S. and Vandenberghe, L., \emph{Convex optimization}, Cambridge university
  press, 2004.

\bibitem[{Douglas and Rachford(1956)}]{douglas56}
Douglas, J. and Rachford, H.~H., \enquote{On the numerical solution of heat
  conduction problems in two and three space variables,} \emph{Transactions of
  the American mathematical Society}, 82(2):421--439, 1956.

\bibitem[{Golub and Pereyra(2003)}]{golub03}
Golub, G. and Pereyra, V., \enquote{Separable nonlinear least squares: the
  variable projection method and its applications,} \emph{Inverse problems},
  19(2):R1, 2003.

\bibitem[{Holland and Welsch(1977)}]{holland77}
Holland, P.~W. and Welsch, R.~E., \enquote{Robust Regression Using Iteratively
  Reweighted Least-Squares,} \emph{Communications in Statistics -- Theory and
  Methods}, 6(9):813--827, 1977.

\bibitem[{Kaufman(1975)}]{kaufman75}
Kaufman, L., \enquote{A variable projection method for solving separable
  nonlinear least squares problems,} \emph{BIT Numerical Mathematics},
  15(1):49--57, 1975.

\bibitem[{Lions and Mercier(1979)}]{lions79}
Lions, P.-L. and Mercier, B., \enquote{Splitting algorithms for the sum of two
  nonlinear operators,} \emph{SIAM Journal on Numerical Analysis},
  16(6):964--979, 1979.

\bibitem[{MacTavish and Barfoot(2015)}]{mactavish_crv15}
MacTavish, K.~A. and Barfoot, T.~D., \enquote{At All Costs: A Comparison of
  Robust Cost Functions for Camera Correspondence Outliers,} in
  \emph{Proceedings of the 12th Conference on Computer and Robot Vision (CRV)},
  pages 62--69, Halifax, Canada, 2015.

\bibitem[{Yang et~al.(2020)Yang, Shi, and Carlone}]{yang20}
Yang, H., Shi, J., and Carlone, L., \enquote{Teaser: Fast and certifiable point
  cloud registration,} \emph{arXiv preprint arXiv:2001.07715}, 2020.

\bibitem[{Zhang(1997)}]{zhang97}
Zhang, Z., \enquote{Parameter Estimation Techniques: A Tutorial with
  Application to Conic Fitting,} \emph{Image and Vision Computing},
  15(1):59--76, 1997.

\end{thebibliography}
